\documentclass{article}
\usepackage[utf8]{inputenc}
\usepackage{amsmath, amssymb}
\usepackage{graphicx}
\usepackage{hyperref}
\hypersetup{
  colorlinks=true,
  linkcolor=black,
  citecolor=blue,
  urlcolor=blue
}
\usepackage{geometry}
\title{Information Gravity: A Field-Theoretic Model for Token Selection in Large Language Models}
\author{Maryna Vyshnyvetska}
\date{}

\begin{document}

\maketitle

\begin{abstract}
We propose a theoretical model called "information gravity" to describe the text generation process in large language models (LLMs). The model uses physical apparatus from field theory and spacetime geometry to formalize the interaction between user queries and the probability distribution of generated tokens. A query is viewed as an object with "information mass" that curves the semantic space of the model, creating gravitational potential wells that "attract" tokens during generation. This model offers a mechanism to explain several observed phenomena in LLM behavior, including hallucinations (emerging from low-density semantic voids), sensitivity to query formulation (due to semantic field curvature changes), and the influence of sampling temperature on output diversity.
\end{abstract}

\section{Introduction}
Large language models (LLMs) have revolutionized the field of artificial intelligence, demonstrating text understanding and generation capabilities approaching human levels. However, despite impressive results, the internal functioning mechanisms of these models largely remain a ``black box.'' As Amodei~\cite{amodei2024urgency} notes in his essay ``The Urgency of Interpretability,'' researchers have limited understanding of why LLMs generate specific responses and how they arrive at their conclusions. This lack of transparency becomes increasingly problematic as LLMs begin to play central roles in economics, technology, and national security.

Of particular concern are phenomena such as unpredictable hallucinations, extreme sensitivity to query formulations, and puzzling patterns in the probability distributions of generated tokens. These phenomena not only limit the reliability of LLMs in critical applications but also point to fundamental gaps in our understanding of their operation.

\subsection{Interpretability Challenges in LLMs}

Contemporary approaches to interpreting LLMs focus on various aspects of their functioning: from mechanistic interpretability of individual components~\cite{olah2020mechanistic} to analyses of attention and neuron activations~\cite{elhage2021attention}. However, most of these approaches remain fragmentary and do not offer a unified explanatory structure for understanding model behavior as a complete system.

Research shows that even minimal changes in query formulation can lead to radically different responses~\cite{cox2025uncertainty}. Sclar et al.~\cite{sclar2023sensitivity} discovered quality variations of up to 56 percentage points with different query formats for the same task. Such sensitivity indicates the existence of complex internal dynamics that cannot be explained by simple statistical approximation.

Another problem relates to ``unnatural'' probability distributions. Zhang et al.~\cite{zhang2024diffuse} demonstrated that when asked for random numbers or names, models show strong preferences for certain responses, although a uniform distribution would be expected as the natural intuitive model of uncertainty for random number or name requests. This suggests the presence of hidden ``attractors'' in the generation process.

\subsection{The Need for a Unified Theoretical Model}

To overcome these issues, a theoretical model is needed that could:
\begin{enumerate}
    \item Integrate disparate observations about LLM behavior into a unified explanatory structure
    \item Propose a mechanism for understanding and predicting hallucinations
    \item Explain model sensitivity to query formulations
    \item Present a conceptual framework for managing the generation process
\end{enumerate}

Existing approaches, such as Energy-Based Models, information geometry, and the semantic landscape paradigm~\cite{gokhale2023semantic}, offer individual elements of such a theory but do not integrate them into a cohesive model.

\subsection{Proposed Approach: Information Gravity}

In this paper, we present a new theoretical model called ``information gravity,'' which draws an analogy between physical gravity and text generation processes in LLMs. According to this model, a user query can be viewed as an object with information ``mass'' that curves the model's semantic space, creating ``gravitational wells'' that attract the generation process toward certain tokens.

The key element of our approach is the formalization of the following concepts:
\begin{enumerate}
    \item Information mass of a query as a function of its entropy, context depth, and novelty
    \item Semantic potential as a measure of token selection probability
    \item Information gravity as the gradient of this potential
    \item Thermodynamic interpretation of the generation process
\end{enumerate}

This model allows us not only to explain observed phenomena in LLM operation but also to propose specific methods for improving their functioning, including information mass management, adaptive sampling temperature tuning, and the development of new quality metrics.

\subsection{Paper Structure}

In the following sections, we elaborate on the proposed information gravity model. Section 2 formalizes the key concepts of the model. Section 3 describes the dynamics of token selection in terms of movement through curved semantic space. Section 4 explains how the model predicts hallucinations, query sensitivity, and temperature effects. Section 5 proposes experimental approaches for theory verification. Finally, section 6 discusses practical applications and perspectives for model development.

Through the lens of information gravity, we aim not only to gain a deeper understanding of LLMs' internal mechanisms but also to expand our conceptual toolkit for working with these increasingly important artificial intelligence systems.

\section{The Concept of Information Gravity}
In large language models (LLMs), the process of generating a response to a user query can be viewed as the model's movement through a specific probabilistic space consisting of tokens and their combinations. Similar to how physical mass distorts spacetime, creating gravitational fields, the information in a prompt distorts the LLM's semantic space, forming areas of ``information attraction.'' This ``information gravitational field'' determines the selection of subsequent tokens during generation, making token generation not a random choice but a directed movement across the semantic landscape.

The idea of viewing information as a source of attraction has parallels in various fields. In physics, the Informational Gravity Hypothesis (IGH)~\cite{chang2022igh} describes gravity as an emergent phenomenon generated by information processes. In machine learning, the cognitive gravitation model~\cite{tang2016cognitive} for data classification uses self-information of an object as an analog of mass, defining the force of attraction between samples proportionally to their information content. In cognitive science, intelligence is sometimes compared to plasma, whose movement is dictated by information gravity~\cite{lee2018possible}.

\subsection{Latent Space of the Model}

We define the latent space of the model L as an n-dimensional space of hidden representations, where each token, context, or attention vector is represented as a point or region. The dimensionality of this space is determined by the model's architecture and corresponds to the dimensionality of its internal representations.

Within this space, the model performs computations, and this is precisely where the ``curvature'' we discuss occurs. This is not an arbitrary metaphor --- transformers truly transform input data through a series of non-linear projections, creating complex geometry in the probability space, which aligns with the Semantic Landscape Paradigm proposed by Gokhale~\cite{gokhale2023semantic}.

It's important to note that the attention mechanism directly influences the formation of this latent space. Attention scores function as analogs of field lines, directing the model toward specific areas of this space and thus forming concrete generation paths. Therefore, attention can be explicitly formalized as a set of vectors defining directions of movement across the semantic landscape, thereby creating ``gravitational channels'' through which token generation flows. This aligns with research by Dar et al.~\cite{dar2023positionaware}, who showed that transformer parameters can be directly interpreted in the vocabulary embedding space.

Additionally, attention scores in each attention layer of the transformer can be interpreted as a matrix that defines not only direction but also the relative strength of interaction between tokens. Thus, the attention matrix serves as an analog of a tensor describing the intensity of the information field, further strengthening the analogy with physical field theories.

\subsection{Information Mass of a Query}

The information mass of a query reflects its ability to distort semantic space and form areas of probabilistic attraction. We propose to formalize this mass as a weighted combination of three main parameters:

\begin{equation}
M(Q) = \alpha \cdot H(Q) + \beta \cdot D(Q) + \gamma \cdot N(Q)
\end{equation}

where:
\begin{itemize}
    \item $H(Q)$ is the information entropy of the query, characterizing uncertainty and ambiguity.
    \item $D(Q)$ is the context depth, reflecting the degree of connection between the query and previous parts of the dialogue or model knowledge.
    \item $N(Q)$ is the novelty of the query, assessing the distance between the current query and the distribution of queries on which the model was trained.
    \item $\alpha, \beta, \gamma$ are coefficients that allow adjusting the model for specific tasks and data types.
\end{itemize}

The components of information mass can be measured independently through the following approaches:
\begin{itemize}
    \item Entropy $H(Q)$: standard Shannon entropy, calculated based on token probabilities.
    \item Context depth $D(Q)$: evaluation through total mutual information, measured using a language model or through embedding metrics (e.g., cosine similarity between query and context embeddings).
    \item Novelty $N(Q)$: Kullback-Leibler divergence between the empirical distribution of query tokens and previously collected statistics from training data.
\end{itemize}

\subsubsection{Information Entropy of the Query}

The information entropy of a query $H(Q)$ can be calculated as the Shannon entropy of the probability distribution of tokens comprising the query:

\begin{equation}
H(Q) = -\sum_{t \in Q} P(t|Q)\log P(t|Q)
\end{equation}

where $P(t|Q)$ is the probability of token $t$ in the context of query $Q$.

High entropy indicates semantic ambiguity or complexity of the query. The higher the entropy, the stronger the curvature of the semantic space and the more difficult it is for the model to determine the ``right'' generation path.

\subsubsection{Context Depth}

Context depth $D(Q)$ reflects the degree of integration of the query into the previous context and can be formalized as the sum of mutual information between the query and previous parts of the dialogue:

\begin{equation}
D(Q) = \sum_{i=1}^{k} I(Q;C_i)
\end{equation}

where $I(Q;C_i)$ is the mutual information between query $Q$ and the $i$-th part of the previous context $C_i$, and $k$ is the depth of the considered context.

High context depth creates complex, extended ``gravitational fields'' that may have multiple local minima, increasing the probability of the model being ``captured'' in suboptimal areas of the semantic space.

\subsubsection{Query Novelty}

The novelty of a query $N(Q)$ represents a measure of how different the query is from the distribution of data on which the model was trained:

\begin{equation}
N(Q) = D_{KL}(P(Q)||P_{train}(Q))
\end{equation}

where $D_{KL}$ is the Kullback-Leibler divergence between the probability distribution of the current query $P(Q)$ and the distribution in the training data $P_{train}(Q)$.

High novelty means that the model has few ``reference points'' in its knowledge space for processing the query, which increases uncertainty and raises the risk of hallucinations.

\subsection{Semantic Field and Its Potential}

We view the space of possible tokens as a multidimensional semantic probabilistic field. Each token in this space is characterized by a specific semantic potential, analogous to the gravitational potential in physics. The user's query alters the configuration of this field, creating zones of lowered semantic potential toward which the model gravitates during token generation.

Although the original token space is discrete, the model uses embeddings, which are continuous vectors. It is in this continuous embedding space that the semantic field and corresponding gravitational potentials are formed, which resolves the contradiction between token discreteness and continuity of the gravitational analogy.

The semantic potential for token $t$ with query $Q$ is defined by the formula:

\begin{equation}
\Phi(t, Q) = -\log P(t|Q)
\end{equation}

where $P(t|Q)$ is the probability of selecting token $t$ given query $Q$.

Thus, the potential is inversely proportional to the logarithm of the token's probability: the more probable the token, the lower its potential. In this model, during generation, the model ``gravitates'' toward tokens with minimal potential, similar to how physical bodies move toward the minimum of gravitational potential.

\begin{figure}[htbp]
  \centering
  \includegraphics[width=0.85\textwidth]{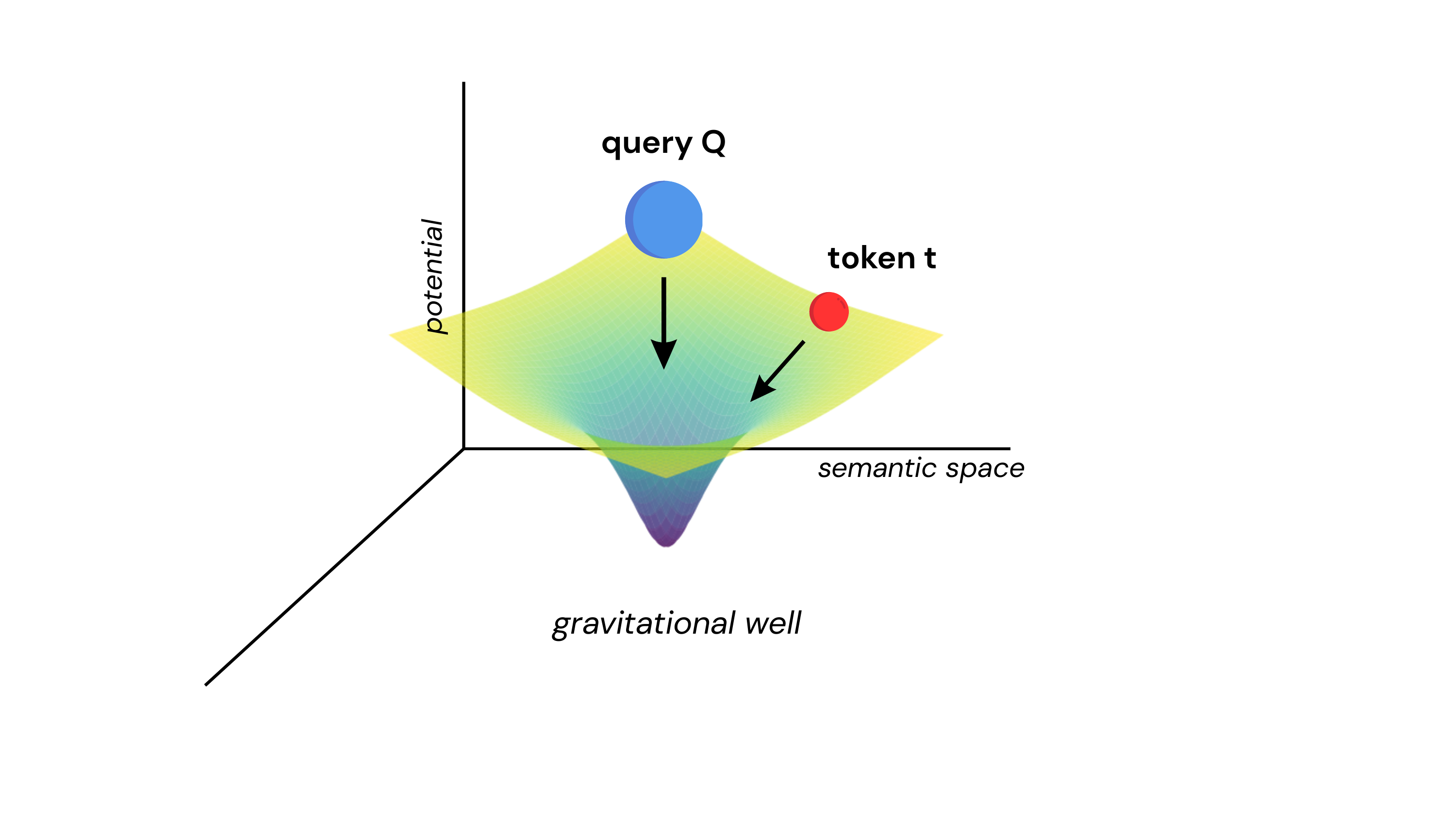}
  \caption{Illustration of information gravity in semantic space. A user query creates a gravitational well in the semantic field, attracting token generation toward areas of minimal semantic potential.}
  \label{fig:gravity}
\end{figure}

\subsection{Information Gravity and Its Gradient}

Based on the semantic potential, we can define information gravity $g(t)$ as the gradient of this potential:

\begin{equation}
g(t) = -\nabla\Phi(t)
\end{equation}

Information gravity indicates the direction of the ``force of attraction'' in semantic space --- the direction in which the model is most likely to move during generation. The higher $g(t)$, the stronger the ``attraction'' to the corresponding token.

It's worth noting that transformers consist of several attention layers, each of which forms its own semantic subspace. As a result, the real structure of the model's information field represents a complex multi-level system of nested ``gravitational fields'' interacting with each other and jointly determining the final probability distribution of tokens.

\section{Dynamics of Token Selection}
In the process of text generation, an LLM selects the next token based on a probability distribution that can be expressed through semantic potential and temperature:

\begin{equation}
P(t|Q) = \frac{\exp(-\Phi(t)/T)}{Z}
\end{equation}

where $T$ is the sampling temperature, and $Z$ is the normalizing coefficient (partition function), defined as:

\begin{equation}
Z = \sum_{t} \exp(-\Phi(t)/T)
\end{equation}

This formula takes the form of a Boltzmann distribution in statistical physics, where temperature $T$ regulates the ``thermal motion'' in the system. At low temperature $(T \rightarrow 0)$, the model rigidly selects the token with minimum potential, while at high temperature $(T \rightarrow \infty)$, the selection becomes more equiprobable, corresponding to high entropy in the output distribution.

Such a thermodynamic interpretation of neural networks is found in work on Energy-Based Models (EBM), and Xie et al.~\cite{xie2021ebm} showed that the learning process can be interpreted as a dynamic alternating projection in the space of distributions in terms of information geometry.

An interesting analogy is the comparison of high-entropy generation with quantum uncertainty, where the model simultaneously considers several possible generation paths, leading to stochastic and less predictable behavior. This opens possibilities for further research into analogies between the thermodynamic behavior of LLMs and quantum systems, especially in the context of multi-agent or ensemble approaches to generation.

\subsection{Generation Trajectory as Movement in a Gravitational Field}

The text generation process can be viewed as movement along a trajectory of minimal action in curved semantic space. At each generation step, the model selects the next token guided by the configuration of the semantic potential created by the query and the previous context.

It is important to note that each selected token modifies the context and, consequently, changes the configuration of the semantic field for subsequent steps. This can be visualized as a moving source of gravity, constantly restructuring the potential field around itself.

The model demonstrates consistency in the style and theme of responses, which can be interpreted as information inertia --- a tendency to maintain the direction of generation after the model has begun movement along a certain path in semantic space. This phenomenon is analogous to inertia in physics and is explained by the fact that already selected tokens create additional information attraction in the given direction.

Formally, the generation trajectory $\tau$ represents a sequence of tokens $t_1, t_2, \ldots, t_n$ that minimizes the action integral:

\begin{equation}
S[\tau] = \int_{\tau}\Phi(t(\tau), Q)d\tau
\end{equation}

taking into account temperature fluctuations that may ``eject'' the system from a local minimum.

\section{Effects and Predictions of the Model}
The proposed information gravity model explains several phenomena observed in LLM operation:

\subsection{Mechanism of Hallucinations}

Hallucinations in LLMs can be explained through the concept of semantic voids—regions of the semantic field with sparse or absent connections to the training data. When the information mass of a query $M(Q)$ is high, but the model's local knowledge is insufficient, the system encounters such voids. Unable to find a strongly connected continuation, the model is forced to select a weakly associated or isolated token, resulting in semantically inconsistent or factually incorrect content.

Mathematically, this can be expressed through an instability condition:

If the Hessian of the semantic potential (matrix of second partial derivatives)
\begin{equation}
H_{ij} = \frac{\partial^2\Phi}{\partial x_i\partial x_j}
\end{equation}
has high eigenvalues in small regions of the latent space, and the global probability of the token $P(t) \ll 1$, then the system is prone to ``capture'' in a false local minimum, leading to the generation of plausible but factually incorrect content.

This mechanism explains why hallucinations often occur with complex, ambiguous queries, or when the model encounters new, unknown information --- in these cases, the semantic field has a complex, ``rugged'' structure with multiple local minima.

An interesting phenomenon is the emergence of so-called information black holes — artificially induced regions of semantic space with extremely high density and low entropy, toward which the model is strongly attracted due to external constraints (such as alignment training, safety instructions, or repetitive prompts). In these areas, the model tends to generate highly predictable, repetitive, or predefined responses.

On the other hand, hallucinations occur when the model encounters semantic voids—regions with sparse or absent connections to the training data. In such cases, the model is forced to select the nearest weakly connected token, resulting in semantically inconsistent or factually incorrect content. Unlike black holes, voids represent areas of low semantic density and weak connectivity.

\subsection{Sensitivity to Query Structure}

The information gravity model predicts that small changes in query formulation can significantly alter the configuration of the semantic field. This explains the empirically observed high sensitivity of LLMs to the format and structure of queries, described by Cox et al.~\cite{cox2025uncertainty} as prompt sensitivity.

This effect can be quantitatively assessed through changes in the potential landscape during query perturbation:
\begin{equation}
\Delta\Phi(t) = \Phi(t, Q') - \Phi(t, Q)
\end{equation}
where $Q'$ is a modified version of query $Q$.

The larger $|\Delta\Phi|$, the more strongly the response probability distribution changes with minimal query modification, which can lead to sharp transitions between different ``attractors'' in semantic space. Research by Sclar et al.~\cite{sclar2023sensitivity} confirms this, demonstrating quality variations of up to 56 percentage points with different query formats.

\subsection{Temperature Effects and Generation Entropy}

The temperature parameter $T$ in the model regulates the balance between ``gravitational attraction'' to local potential minima and ``thermal fluctuations'' that allow exploration of broader regions of semantic space.

At low temperatures $(T \rightarrow 0)$, the model strictly follows the potential gradient, selecting the most probable tokens, leading to deterministic but potentially stereotypical generation. At high temperatures $(T \rightarrow \infty)$, the influence of the ``gravitational field'' weakens, and generation becomes more random, which can lead to more creative but also more chaotic results.

This effect corresponds to observations by Zhang et al.~\cite{zhang2024diffuse}, who found that LLMs demonstrate insufficient diffuseness in distribution when generating stochastic content, preferring certain tokens with disproportionately high probability. For example, when asked to output a random number between 1 and 10, the Llama-2-13B-chat model unevenly prefers the number 5, and when generating a random name, the Mistral-7B model selects ``Avery'' 40 times more frequently than expected.

Quantitatively, the entropy of the output distribution increases with temperature:
\begin{equation}
\lim_{T \rightarrow 0} H(P(t|Q, T)) = 0 \quad \lim_{T \rightarrow \infty} H(P(t|Q, T)) = \log(|V|)
\end{equation}
where $|V|$ is the vocabulary size of the model.

\subsection{Model Limitations}

Despite its theoretical appeal, the information gravity model has several limitations:
\begin{itemize}
    \item Empirical verification requires complex experiments and large amounts of data.
    \item Measuring the components of information mass is complicated by their internal interrelationship.
    \item The complexity of visualizing and interpreting multidimensional semantic space, especially in large models.
    \item The analogy with physics, while intuitively clear, may risk misinterpretation if taken as a literal physical model rather than a metaphorical mapping and should be interpreted as a conceptual framework rather than a direct physical analogy.
\end{itemize}

\section{Experimental Approaches}
Experiments for measuring the components of information mass should be conducted separately and independently from each other. For example, for entropy---formulate queries with varying degrees of ambiguity (synonyms, polysemy); for context depth---vary the length and complexity of previous messages; for novelty---use queries with controlled distance from the training corpus (e.g., combining well-known and entirely new concepts). Then experimentally verify the influence of each factor on generation behavior.

For empirical validation of the information gravity theory, we propose the following experimental approaches:

\subsection{Visualization of the Semantic Landscape}

One of the key aspects of experimental verification is the visualization of the semantic potential $\Phi(t)$ and observation of its changes with different queries. This can be implemented using dimensionality reduction methods, such as $t-SNE$ or $UMAP$, applied to the model's logit vectors before applying softmax:

\begin{enumerate}
    \item For a set of different queries $Q_1, Q_2, \ldots, Q_m$, compute logit vectors $L_i$, i.e., the raw outputs of the model's final linear transformation before the softmax layer.
    \item Apply $t-SNE$ or $UMAP$ to the obtained vectors for projection into two-dimensional or three-dimensional space.
    \item Visualize the resulting projections as ``potential maps,'' where darker areas correspond to lower potential (higher probability).
\end{enumerate}

This approach aligns with methods proposed for visualizing semantic landscapes in the work of Gokhale~\cite{gokhale2023semantic}, and can be complemented by methods developed by Pingbang Hu and Mahito Sugiyama~\cite{hu2024projection} for interpretable data generation using energy-based models.

Such an approach will visually demonstrate how different queries ``deform'' the semantic space, creating various configurations of ``gravitational wells.''

\subsection{Analysis of the Relationship Between Information Mass and Hallucinations}

To confirm the hypothesis about the relationship between query information mass and the model's tendency toward hallucinations, the following experiment is proposed:

\begin{enumerate}
    \item Create a set of queries with controlled increases in information mass components: entropy $H(Q)$, context depth $D(Q)$, and novelty $N(Q)$.
    \item For each query, generate multiple responses with varying temperature T.
    \item Assess the frequency and severity of hallucinations in the generated responses.
    \item Analyze the correlation between information mass components and hallucination metrics.
\end{enumerate}

Model prediction: queries with high entropy, deep context, and high novelty will demonstrate significantly higher hallucination frequency, especially at low temperatures, where the model does not have the ability to ``explore'' alternative paths in semantic space.

\subsection{Investigation of Temperature Effects}

To study the influence of temperature on semantic field configuration, we propose:

\begin{enumerate}
    \item Select a set of fixed queries of varying complexity.
    \item For each query, generate multiple responses with systematically varying temperature parameter $T$ in a range from near 0 to very high values.
    \item For each value of $T$, analyze:
    \begin{itemize}
        \item Shannon entropy of the token probability distribution
        \item Diversity of generated responses
        \item Semantic coherence
        \item Hallucination frequency
    \end{itemize}
\end{enumerate}

Model prediction: at low temperatures, ``gravitational capture'' will be observed---the model will consistently select the same tokens, following the potential gradient. At high temperatures, ``thermal motion'' will dominate over ``gravitational attraction,'' and token selection will become more random, increasing diversity but potentially reducing coherence.

\section{Perspectives and Applications}
The concept of information gravity opens broad perspectives for practical use and further research:

\subsection{Improving Generation Through Information Mass Management}

Understanding how information mass components influence the curvature of semantic space, we can develop methods for more precise control over generation:

\begin{enumerate}
    \item \textbf{Entropy Balancing:} Modifying queries to achieve an optimal level of entropy $H(Q)$ that is sufficient for creative generation but insufficient to provoke hallucinations.
    
    \item \textbf{Context Stabilization:} Managing context depth $D(Q)$ to create more stable ``gravitational fields'' with fewer local minima.
    
    \item \textbf{Novelty and Anchoring:} Compensating for high query novelty $N(Q)$ by adding ``anchor'' fragments that connect the query with information known to the model, reducing the risk of ``gravitational collapse'' in incorrect areas of semantic space.
\end{enumerate}

These methods may be particularly useful in light of observations by Cox et al.~\cite{cox2025uncertainty} regarding high LLM sensitivity to query formulation, and may complement approaches proposed by Zhang et al.~\cite{zhang2024diffuse} for achieving more diffuse distribution during generation.

\subsection{Adaptive Temperature Control}

The information gravity model provides a theoretical foundation for developing algorithms for adaptive sampling temperature control based on query properties:

To modulate generation behavior in relation to semantic field complexity, we propose the following formulation:

\begin{equation}
T_{adaptive}(Q) = f(M(Q))
\end{equation}

where function $f$ maps the information mass of the query to an optimal temperature value.

Queries with high information mass may require higher temperatures to overcome ``gravitational traps'' and explore broader areas of semantic space, while simple, unambiguous queries can be efficiently processed at low temperatures.

\subsection{Quality and Interpretability Metrics for LLMs}

The concept of information gravity can be used to develop new metrics for evaluating LLM quality and interpretability:

We propose the following novel metrics for evaluating LLM performance:
\begin{itemize}
    \item \textbf{Gravitational Curvature:} a metric assessing the degree of heterogeneity in the semantic field (e.g., through variations in potential gradient). High curvature indicates instability and tendency toward hallucinations.
    
    \item \textbf{Information Conductivity:} a metric evaluating the efficiency of semantic information transfer from query to response. This can be measured through mutual information or cross-entropy between query and response.
    
    \item \textbf{Semantic Field Stability:} Assessment of semantic potential configuration stability against small query variations as a measure of model robustness.
\end{itemize}

These metrics align with the research direction proposed in the work of Pingbang Hu and Mahito Sugiyama~\cite{hu2024projection} and can be integrated with information geometry approaches described by Xie et al.~\cite{xie2021ebm}.

Such metrics can provide deeper understanding of internal processes in LLMs and help in developing more reliable and interpretable next-generation models.

\section{Conclusion}
In this paper, we have presented a theoretical model of information gravity for explaining and predicting the behavior of large language models in the process of text generation. The proposed model creates a conceptual bridge between physical theories, information geometry, and probabilistic processes in LLMs, offering a unified paradigm for analyzing the behavior of these systems.

\subsection{Key Results}

We have formalized the following key concepts:
\begin{enumerate}
    \item Information mass of a query as a combination of entropy, context depth, and novelty, influencing the ``curvature'' of the model's semantic space.
    
    \item Semantic potential as a metric describing the probabilistic landscape of text generation, analogous to gravitational potential in physics.
    
    \item Information gravity as the gradient of semantic potential, governing the trajectory of token selection within the semantic landscape.
    
    \item Thermodynamic interpretation of the generation process, where sampling temperature regulates the balance between ``gravitational attraction'' and ``thermal fluctuations''.
\end{enumerate}

This theoretical framework explains several empirically observed phenomena in LLM operation:
\begin{itemize}
    \item The mechanism of hallucination emergence through semantic voids—regions of semantic space with sparse or absent connections to the training data, resulting in semantically inconsistent or factually incorrect content.
    
    \item High model sensitivity to query formulation as changes in semantic field configuration
    
    \item The influence of temperature on generation diversity and quality through changes in the system's thermodynamic regime
    
    \item The existence of artificially induced information black holes, created by alignment instructions, adversarial prompts, or repetitive query patterns, leading to highly predictable or constrained model outputs
    
    \item Information inertia, explaining stylistic and thematic consistency in generation
\end{itemize}

\subsection{Perspectives for Model Development}

The proposed information gravity model opens several promising directions for further research:
\begin{enumerate}
    \item Empirical verification of the theory through visualization of semantic landscapes, analysis of hallucinations, and investigation of temperature effects.
    
    \item Practical applications in the form of new methods for controlling generation through information mass manipulation, adaptive temperature control, and development of new model quality metrics.
    
    \item Theoretical development of the model through formalization of connections with attention mechanisms in transformers, integration with information geometry, and extension to multimodal models.
    
    \item Inter-model interaction --- studying ``information fields'' created by different models and their interactions in the context of collaborative work among multiple LLMs.
\end{enumerate}

\subsection{Possible Limitations}

Despite the explanatory power of the proposed model, it has several potential limitations:
\begin{itemize}
    \item Difficulty in independently measuring information mass components
    
    \item Challenges in formalizing complex interaction of information fields between different attention layers in multilayer transformers
    
    \item Simplified representation of semantic space continuity given token discreteness
\end{itemize}

These limitations, however, represent directions for further model development rather than fundamental obstacles to its applicability.

\subsection{Concluding Remarks}

The theory of information gravity offers not just a new metaphor for understanding processes in LLMs, but also a potentially verifiable scientific hypothesis with practical applications. As large language models become increasingly powerful and pervasive tools, understanding the fundamental principles of their operation takes on critical importance.

The proposed model can serve as a step toward deeper understanding and control over LLM behavior, providing a theoretical foundation for developing more interpretable, reliable, and manageable artificial intelligence systems. In light of recent discussions about ``black boxes'' in machine learning, the presented theory of information gravity may serve as a conceptual instrument for improving interpretability of these black-box systems and enabling more transparent operation of language models.

\section*{Acknowledgements}
With illogical tenderness to ChatGPT and Claude for their role in preparing the manuscript.

This work is also available as a Zenodo preprint: \url{https://doi.org/10.5281/zenodo.15283647}

\bibliographystyle{plain}
\bibliography{references}

\begin{thebibliography}{10}

\bibitem{amodei2024urgency}
Dario Amodei.
\newblock The urgency of interpretability, 2024.
\newblock Unpublished manuscript.

\bibitem{chang2022igh}
J.~Chang.
\newblock The informational gravity hypothesis.
\newblock {\em OSF Preprints}, 2022.

\bibitem{cox2025uncertainty}
J.~Cox, Y.~Han, D.~Wang, S.~Nair, and Z.~Yu.
\newblock Understanding uncertainty in large language models through prompt perturbations.
\newblock In {\em AAAI 2025}, 2025.

\bibitem{dar2023positionaware}
Y.~Dar, Z.~Ren, L.~Moysis, S.~Agarwal, T.~Zhang, and V.~J. Hellendoorn.
\newblock Understanding embeddings in the transformer through the lens of the position-aware linear model.
\newblock {\em arXiv preprint}, 2023.

\bibitem{elhage2021attention}
Nelson Elhage et~al.
\newblock Analyses of attention and neuron activations, 2021.
\newblock Technical report.

\bibitem{gokhale2023semantic}
C.~Gokhale.
\newblock The semantic landscape paradigm for neural networks.
\newblock {\em ResearchGate}, 2023.

\bibitem{hu2024projection}
P.~Hu and M.~Sugiyama.
\newblock Interpretable data generation from energy-based models by projection.
\newblock {\em arXiv preprint}, 2024.

\bibitem{lee2018possible}
J.~Lee.
\newblock Is artificial intelligence possible?
\newblock {\em arXiv preprint}, 2018.

\bibitem{olah2020mechanistic}
Chris Olah et~al.
\newblock Mechanistic interpretability of individual components, 2020.
\newblock Research blog.

\bibitem{sclar2023sensitivity}
M.~Sclar, P.~West, P.~Liang, and N.~Saunshi.
\newblock Quantifying the sensitivity of large language models to different prompt formats.
\newblock {\em arXiv preprint}, 2023.

\bibitem{tang2016cognitive}
M.~Tang and X.~Yang.
\newblock Classification using cognitive gravitation model.
\newblock In {\em Lecture Notes in Computer Science}, volume 10016, pages 203--214, 2016.

\bibitem{xie2021ebm}
J.~Xie, K.~C. Li, Y.~Zhu, and C.~H. Zhang.
\newblock Cooperation in variational and non-variational mcmc via energy-based models.
\newblock In {\em AAAI}, 2021.

\bibitem{zhang2024diffuse}
L.~Zhang, W.~Chen, W.~Chen, A.~Wang, Q.~Deng, Y.~Tao, and C.~Finn.
\newblock Tuning language models to generate diffuse distributions.
\newblock {\em arXiv preprint}, 2024.

\end{thebibliography}

\end{document}